\documentclass[conference]{ieeeconf}  % Comment this line out if you need a4paper

\IEEEoverridecommandlockouts                              % This command is only needed if 
\usepackage{cite}
\usepackage{xcolor,soul}
\usepackage{tcolorbox} 
\soulregister\cite7
\soulregister\citep7
\soulregister\citet7
\soulregister\ref7
\soulregister\pageref7

\usepackage{amssymb}
\usepackage{bm}
\usepackage{flushend}

\usepackage{multirow}
\usepackage{soul}
\usepackage{svg}
\usepackage{float}

\usepackage{hyperref}
\usepackage{booktabs}

\hypersetup{
    colorlinks=true,
    linkcolor=black,
}

\usepackage{cuted}
\usepackage{capt-of}
\usepackage{caption}
\usepackage{amsmath}

\usepackage{blindtext}
\usepackage{tabularx}

\usepackage{amsmath}

\title{\LARGE \bf
LiMo-Calib: On-Site Fast LiDAR-Motor Calibration for Quadruped Robot-Based Panoramic 3D Sensing System}

\author{Jianping Li*,~\IEEEmembership{Member,~IEEE}, Zhongyuan Liu*, Xinhang Xu, Jinxin Liu, \\Shenghai Yuan,~\IEEEmembership{Member,~IEEE}, Xu Fang, and Lihua Xie,~\IEEEmembership{Fellow,~IEEE}}

\begin{document}
% \begin{linenumbers}
% \pagewiselinenumbers 
% \switchlinenumbers
\maketitle

\renewcommand{\thefootnote}{}
\footnotetext{This work was supported by China-Singapore International Joint Research Institute (CSIJRI) Development Plan for the Technology Application Center (TAC) and National Research Foundation, Singapore, under its Medium-Sized Center for Advanced Robotics Technology Innovation. This work was partially supported by open project funding of Key Laboratory of Intelligent Control and Optimization for Industrial Equipment of Ministry of Education under Grant LICO2023YB01. J. Li, Z. Liu, X. Xu, J. Liu, S. Yuan, and L. Xie are with China-Singapore International Joint Research Institute (CSIJRI) and School of Electrical and Electronic Engineering, Nanyang Technological University, 50 Nanyang Avenue, Singapore. X. Fang is with School of Control Science and Engineering, Dalian University of Technology, Dalian 116024, China (E-mail: jianping.li@ntu.edu.sg, zliu051@e.ntu.edu.sg, xu0021ng@e.ntu.edu.sg, jinxin.liu@ntu.edu.sg, shyuan@ntu.edu.sg, fa0001xu@e.ntu.edu.sg, elhxie@ntu.edu.sg) (Jianping Li and Zhongyuan Liu contribute equally to this work.)}

%%%%%%%%%%%%%%%%%%%%%%%%%%%%%%%%%%%%%%%%%%%%%%%%%%%%%%%%%%%%%%%%%%%%%%%%%%%%%%%%

\begin{abstract}

Conventional single LiDAR systems are inherently constrained by their limited field of view (FoV), leading to blind spots and incomplete environmental awareness, particularly on robotic platforms with strict payload limitations. Integrating a motorized LiDAR offers a practical solution by significantly expanding the sensor’s FoV and enabling adaptive panoramic 3D sensing. However, the high-frequency vibrations of the quadruped robot introduce calibration challenges, causing variations in the LiDAR-motor transformation that degrade sensing accuracy. Existing calibration methods that use artificial targets or dense feature extraction lack feasibility for on-site applications and real-time implementation. To overcome these limitations, we propose LiMo-Calib, an efficient on-site calibration method that eliminates the need for external targets by leveraging geometric features directly from raw LiDAR scans. LiMo-Calib optimizes feature selection based on normal distribution to accelerate convergence while maintaining accuracy and incorporates a reweighting mechanism that evaluates local plane fitting quality to enhance robustness. We integrate and validate the proposed method on a motorized LiDAR system mounted on a quadruped robot, demonstrating significant improvements in calibration efficiency and 3D sensing accuracy, making LiMo-Calib well-suited for real-world robotic applications. We further demonstrate the accuracy improvements of the LIO on the panoramic 3D sensing system using the calibrated parameters. The code will be available at: \url{https://github.com/kafeiyin00/LiMo-Calib}.

\end{abstract}

%%%%%%%%%%%%%%%%%%%%%%%%%%%%%%%%%%%%%%%%%%%%%%%%%%%%%%%%%%%%%%%%%%%%%%%%%%%%%%%%
\section{Introduction}

\begin{figure}
    \centering
    \includegraphics[width=\linewidth]{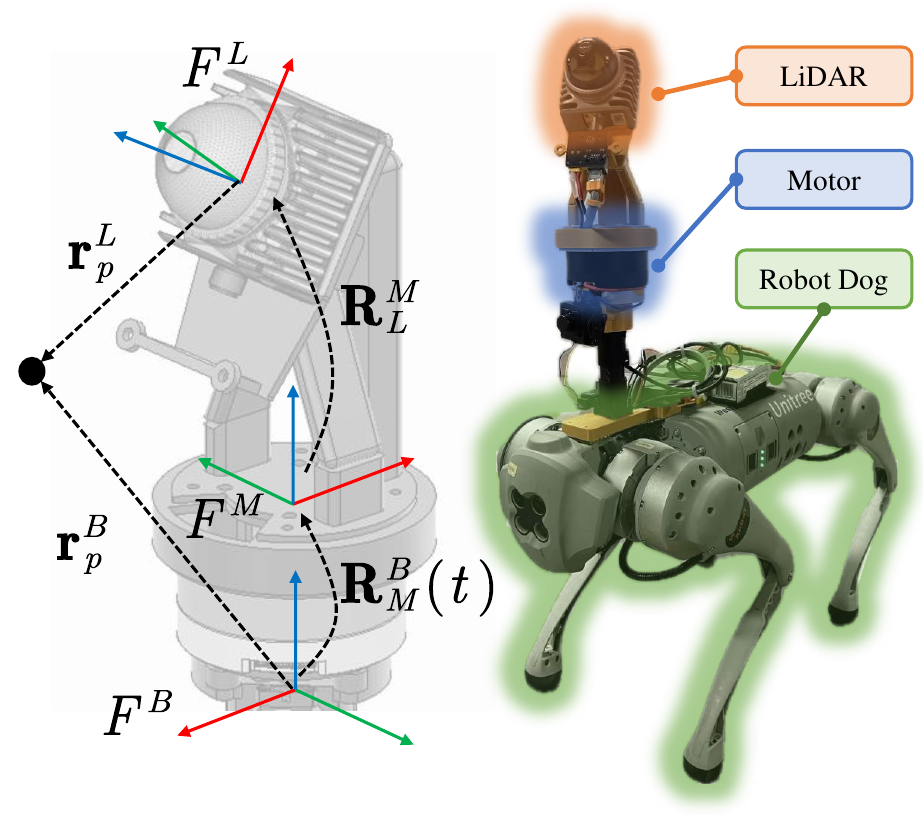}
    \caption{Coordinates and mechanical design of the proposed motorized LiDAR system on a quadruped robot. }
    \label{fig:hardware}
\end{figure}

In recent years, robotic platforms, particularly quadruped robots, have seen significant advancements, making them increasingly valuable for applications such as autonomous inspection, search and rescue, and exploration of complex environments \cite{yang2023iplanner,roth2024viplanner}. However, conventional LiDAR systems are inherently constrained by their limited field of view (FoV), leading to blind spots and incomplete environmental awareness \cite{jiao2021robust}. Given the weight and size constraints of quadruped robots, mounting multiple LiDAR units to enhance FoV is often impractical \cite{hovermap2024,yang2022hierarchical,yuan2021low,zhen2017robust,li2019nrli}. A more efficient solution is integrating a motorized LiDAR \cite{li2024ua} onto the quadruped robot platform for a panoramic 3D sensing system as shown in Fig. \ref{fig:hardware}. By leveraging motor control, the motorized LiDAR substantially expands the sensor’s FoV, enabling broader spatial coverage and adaptive scanning to focus on critical features.

Integrating a motorized LiDAR with a quadruped robot introduces significant calibration issues due to the high-frequency vibrations inherent in the robot’s movement \cite{chai2022survey}. These vibrations and oscillations cause fluctuations in the relative transformation between the LiDAR and the motor, leading to potential misalignment. If not compensated properly on site, these variations can severely degrade the accuracy of 3D sensing, compromising the reliability of the system. 

The calibration of the LiDAR system has been long studied in the field of robotics \cite{alismail2015automatic,kang2016full} and photogrammetry \cite{skaloud2006rigorous,li2023whu}. Artificial targets are usually used for calibration of the LiDAR system \cite{chen2018extrinsic}. But the artificial targets may not be available for the on-site calibration applications. Despite the artificial targets distributed in the calibration filed, planar features extracted from the environment are the most commonly used primitives for automatically constructing the calibration functions \cite{alismail2015automatic,skaloud2006rigorous,liao2023se}. Nevertheless, the large number of extracted planar features significantly increases computational complexity, leading to long processing times that are impractical for edge computing units on mobile robots. Existing LiDAR-motor calibration methods inherently involve a trade-off between the number of extracted primitives and calibration accuracy. A higher number of primitives can enhance accuracy by providing more constraints, but it also leads to increased computational demands and longer processing times \cite{li2024hcto,skaloud2006rigorous,yan2022automatic}. Conversely, reducing the number of primitives improves efficiency but may compromise calibration precision. Achieving an optimal balance between these factors remains a key challenge in developing accurate and efficient on-site calibration solutions for motorized LiDAR systems. To the best of our knowledge, existing methods still struggle to achieve accurate and efficient on-site calibration of motorized LiDAR systems, highlighting the need for a more effective solution.

Addressing these challenges, we propose LiMo-Calib, an on-site fast calibration method that leverages the inherent geometric features in raw LiDAR scans without the need for external targets. To ensure both efficiency and accuracy, we strategically select the most suitable primitives based on their normal distribution, which accelerates optimization while maintaining comparable calibration precision. Additionally, to enhance robustness across diverse scenarios, we introduce a weighted mechanism that evaluates the quality of local plane fitting during the correspondence construction process. This approach prioritizes high-quality features, improving the reliability and adaptability of the calibration method on the quadruped robot. The main contributions of LiMo-Calib are summarized as follows:

(1) LiMo-Calib selects planar primitives based on their normal distribution to accelerate optimization while preserving calibration accuracy.

(2) LiMo-Calib employs a reweighting mechanism to enhance robustness and improve overall accuracy, resulting in precise and reliable panoramic 3D sensing.

(3) LiMo-Calib is evaluated on the in-house LiDAR-motor panoramic 3D sensing system. We further demonstrate the accuracy improvements of the LIO on the panoramic 3D sensing system using the calibrated parameters, which demonstrates efficiency and accuracy in real-world applications.

\section{Related Work}

% Sensor calibration has long been a cornerstone for achieving robust perception in autonomous systems. Early work in multi-sensor calibration—such as LiDAR-IMU and LiDAR–camera calibration—has largely focused on controlled, offline scenarios. Many methods in these areas rely on artificial targets or pre-structured environments to establish accurate correspondences \cite{cui2020,dhall2017}. Although target-based approaches can yield high precision, their dependence on manually deployed calibration objects renders them impractical for on-site applications, particularly in dynamic or unstructured environments.

Sensor calibration is a fundamental requirement for robust perception in autonomous systems. Early efforts in multi-sensor calibration, such as LiDAR-IMU, LiDAR-camera, and LiDAR-motor calibration, predominantly focused on controlled, offline environments. Many established methods in these domains rely on artificial targets or pre-structured settings to establish precise correspondences (e.g., \cite{cui2020, dhall2017,gao2019calibration}). While target-based approaches can achieve high accuracy, their dependence on manually deployed calibration objects limits their practicality for on-site applications, particularly in unstructured environments where such setups are often infeasible.

To overcome the limitations of target-based calibration, recent research has focused on targetless methods that leverage naturally occurring features such as edges, planes, and semantic boundaries from raw sensor data \cite{skaloud2006rigorous,park2020,liao2023se}. These approaches extract dense geometric or semantic features and employ complex nonlinear optimization to align sensor measurements, but they incur substantial computational overhead and are highly sensitive to scene structure; their performance may degrade in environments with sparse salient features or overly complex geometry. Moreover, many of these methods assume static or slowly varying conditions, limiting their real-time applicability on mobile platforms. For instance, Skaloud and Lichti \cite{skaloud2006rigorous} introduced a rigorous self-calibration technique for airborne laser scanning that, despite its foundational role, is not well suited for real-time mobile applications due to high complexity. Park et al. \cite{park2020} proposed a spatiotemporal calibration method for camera–LiDAR systems tailored to dynamic environments, yet it still demands dense feature extraction and intensive nonlinear optimization. In urban settings, Liao et al. \cite{liao2023se} presented SE-Calib, a semantic edge-based approach that enhances robustness through semantic constraints, but its reliance on high-quality semantic segmentation adds further computational burden and sensitivity to lighting variations. Similarly, Yuan et al. \cite{yuan2021} achieved pixel-level extrinsic calibration by aligning natural edge features across LiDAR and camera data, while Liu et al. \cite{liu2021} developed an annotation-free method based on semantic alignment loss; both methods, however, face challenges in computational efficiency and adaptability to dynamic scenes. Collectively, these studies underscore a critical trade-off in targetless calibration between eliminating artificial targets for greater deployment flexibility and managing the increased computational load and environmental sensitivity which motivates our development of LiMo-Calib that judiciously selects high-confidence geometric primitives and incorporates robust reweighting strategies to achieve efficient, accurate calibration in dynamic, on-site applications.

Recent work in rotating LiDAR calibration has largely focused on exploiting planar constraints to determine the sensor’s extrinsic parameters. For instance, Zeng et al. \cite{zeng2018} proposed an improved calibration method that optimizes plane data extracted from raw LiDAR scans to enhance accuracy and robustness; however, its reliance on the presence of prominent planar features and high computational cost limits its applicability in cluttered or sparse environments. Similarly, Olivka et al. \cite{olivka2016} addressed the calibration of short-range 2D laser range finders for 3D SLAM by leveraging environmental planar features. Although effective in structured settings, this approach is sensitive to noise and less suited for dynamic or large-scale applications. Furthermore, Alismail and Browning \cite{alismail2015automatic} proposed an automatic calibration algorithm that estimates spinning LiDAR internal parameters by assuming local planar surfaces, a strategy that fails when the plane assumption does not hold and imposes significant computational burdens. Kang and Doh \cite{kang2016full} presented a full-DOF calibration approach using a simple plane measurement to estimate all six degrees of freedom between a rotating LiDAR and its motor; although it performs well in ideal, structured environments, its accuracy diminishes in unstructured or noisy settings due to its dependence on precise plane extraction. Collectively, while these approaches offer effective calibration solutions for rotating LiDAR systems, they share common limitations in terms of heavy dependence on high computational complexity.

Our work diverges from these prior approaches by addressing the dual challenges of computational efficiency and robustness even in some unstructured environments. Rather than processing dense point clouds or requiring artificial targets, LiMo-Calib extracts reliable planar features from raw LiDAR scans using an adaptive, normal-based selection strategy combined with a reweighting mechanism that evaluates local plane fitting quality. This selective feature extraction not only reduces redundancy and computational overhead but also enhances calibration accuracy under high-frequency vibrations and dynamic motion. In doing so, our method bridges the gap between target-based precision and the practical demands of real-time, on-site calibration.

\section{Hardware and Coordinate Systems}

The proposed panoramic 3D sensing system consists of a quadruped robot integrated with a motorized LiDAR scanner, as illustrated in Fig.~\ref{fig:hardware}. The key components and their corresponding coordinate frames are described as follows:

\textit {Quadruped robot (Body Frame $\{B\}$):} A four-legged mobile robot provides a stable platform for locomotion. All robot navigation and control tasks operate in the robot’s base frame, denoted as $\{B\}$. Internally, the robot contains on-board computing resources that handle sensor data collection and preliminary processing.

\textit {Motor Assembly (Motor Frame $\{M\}$):} On top of the robot, a single-axis motor is mounted vertically. Its rotation axis aligns with the $z$-axis of $\{B\}$. The motor frame, labeled $\{M\}$, thus rotates around $\hat{\mathbf{z}}$ of the base frame at a controlled angular velocity. In our setup, the motor can complete a full $360^\circ$ rotation, enabling panoramic coverage. To support this continuous rotation without entangling cables, a slip ring is placed between the motor’s output shaft and the LiDAR sensor. This slip ring provides uninterrupted electrical connections for both power and data signals, thereby eliminating the risk of twisted wiring during full revolutions. Additionally, the motor’s encoder measures the rotation angle $\theta(t)$ at any given time $t$, providing the fundamental input for synchronization with the LiDAR data.

\textit {LiDAR Sensor (LiDAR Frame $\{L\}$):} We employ a Livox Mid-360 LiDAR, which provides a $360^\circ \times 59^\circ$ field of view (FOV) and a detection range from $0.1\,\mathrm{m}$ up to $40\,\mathrm{m}$ (at $10\%$ reflectivity).

\textit {Degrees of Freedom for the LiDAR ($\{L\}$)-Motor ($\{M\}$) calibration:} In the context of LiDAR-motor calibration, the calibration problem does not strictly involve a full 6-DOF rigid-body transformation because the sensor's motion model is mechanically constrained. For a motorized rotating LiDAR, the model can be described with at most five degrees of freedom: three rotational and two translational. In particular, translations along the LiDAR’s rotation axis (the \(z\)-axis) are unobservable, so only the \(x\)- and \(y\)-axis translations (left–right and front–back) are considered. To further improve calibration accuracy, we also constrain the LiDAR’s yaw angle (rotation around the \(z\)-axis). This is justified by the fact that, physically, the LiDAR rotates horizontally with the motor about the vertical axis, and thus its yaw can be directly obtained from the motor’s angle or is otherwise unnecessary to calibrate. Additionally, the LiDAR’s height (translation along the \(z\)-axis) often has reliable prior information or negligible error and does not need to be optimized. Consequently, we only locally optimize the LiDAR’s roll and pitch with respect to the carrier, as well as the horizontal (\(x\)- and \(y\)-axis) translations, resulting in a final model with four degrees of freedom.

In summary, the proposed hardware design enables a full $360^\circ$ panoramic sweep of the environment by the LiDAR. The robot’s mobility further permits 3D data capture in dynamic or unstructured terrains, making accurate extrinsic calibration $\mathbf{R}_L^M$ and $\mathbf{r}_L^M$ crucial for subsequent tasks such as mapping, navigation, or object detection.

\section{LiMo-Calib}
\label{sec:method}

% Efficient calibration of lidar-motor systems requires reducing the computational burden without sacrificing geometric fidelity. Direct processing of full point clouds often leads to excessive redundancy and increased processing time, which is impractical for field applications. 

\begin{figure} [ht]
    \centering
    \includegraphics[width=\linewidth]{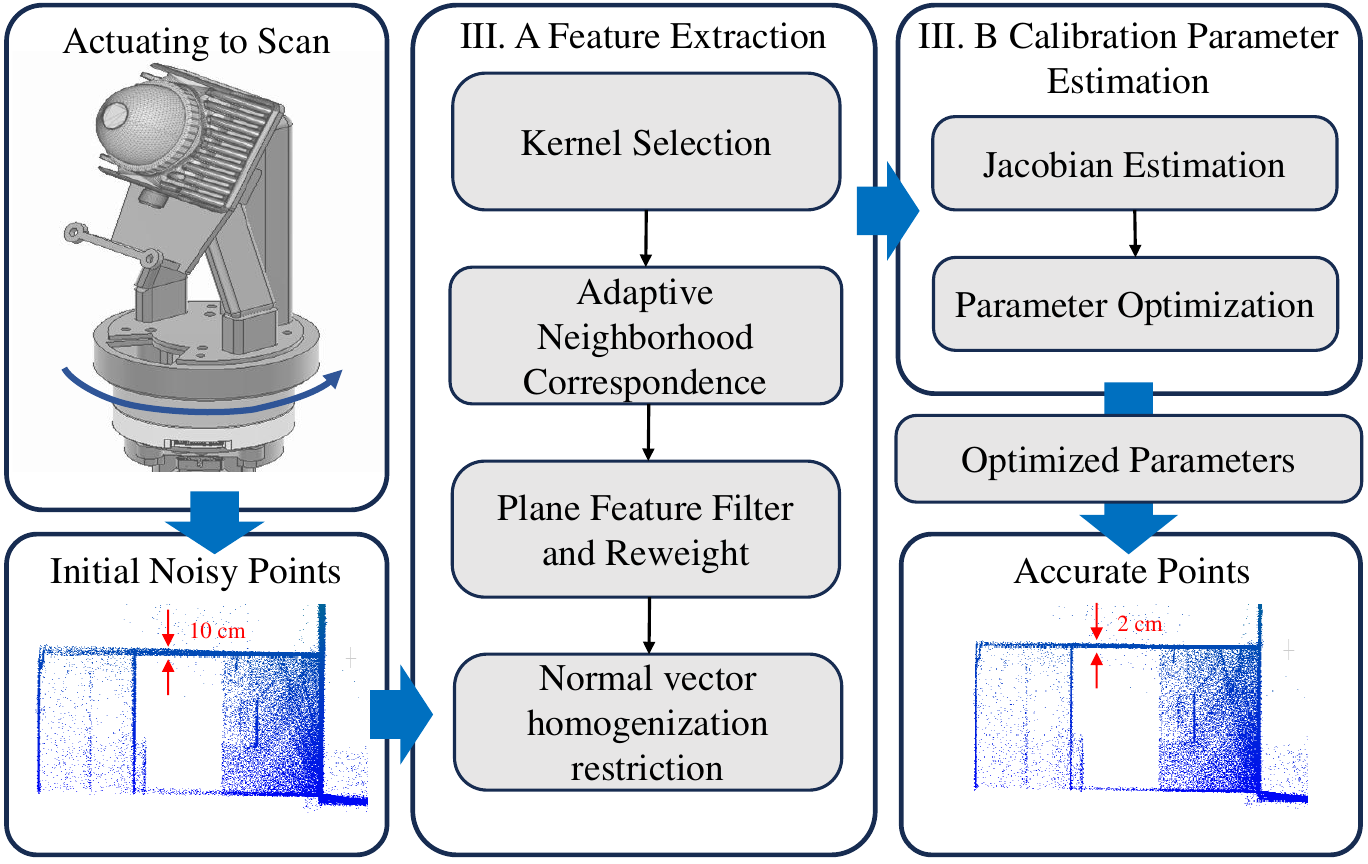}
    \caption{Workflow of the proposed LiMo-Calib, on-site fast LiDAR-motor calibration.}
    \label{fig:workflow}
\end{figure}

To calibrate the LiDAR-motor mounting parameters, the proposed LiMo-calib actuate the motor to get the environmental scanning data first as shown in Fig. \ref{fig:workflow}. Then the calibration process begins by mapping each 3D point $\mathbf{r}_p^L \in \{L\}$ to the base frame $\{B\}$ to get the initial noisy point clouds using the transformation:
\begin{equation}
    \mathbf{r}_p^B(t) \;=\; \mathbf{R}_M^B(t)\,\bigl(\mathbf{R}_L^M\,\mathbf{r}_p^L \;+\; \mathbf{r}_L^M\bigr),
    \label{eq:point_transform}
\end{equation}
where $\mathbf{R}_L^M$ and $\mathbf{r}_L^M$ denote the extrinsic parameters between the LiDAR and the motor frames, and $\mathbf{R}_M^B(t)$ represents the time-varying rotation of the motor relative to the base.

To formulate the calibration objective, let $\mathbf{r}_p^B(t)$ be the coordinates of a point $\mathbf{r}_p^L$ (in the LiDAR frame) expressed in the base frame $\{B\}$ at time $t$, following Eq. \eqref{eq:point_transform}. Suppose we have a set of planar correspondences, each characterized by a plane normal $\mathbf{n}_i$ and two timestamps $t_n$ and $t_m$ such that 3D point $\mathbf{r}_{p_i}^L$ on the same plane correspondence is observed at different motor angles. The corresponding world-frame points are $\mathbf{r}_{p_i}^B(t_n)$ and $\mathbf{r}_{p_i}^B(t_m)$, respectively. We define the residual for each correspondence as the projection of the point displacement onto the plane normal:
\begin{equation}\label{eq:plane_factor_residual}
    r_i \;=\; \mathbf{n}_i^\top\bigl[\mathbf{r}_{p_i}^B(t_m) \;-\; \mathbf{r}_{p_i}^B(t_n)\bigr],
\end{equation}
where $\mathbf{n}_i$ is estimated from local plane fitting around $\mathbf{r}_{p_i}^B(t_n)$ or $\mathbf{r}_{p_i}^B(t_m)$. 

To obtain the calibration parameters $\mathbf{R}_L^M$ and $\mathbf{r}_L^M$, we minimize the weighted sum of squared residuals:
\begin{equation}\label{eq:loss_function}
    \min_{\mathbf{R}_L^M, \,\mathbf{r}_L^M} \;\; \sum_{i} w_i \,\Bigl(\mathbf{n}_i^\top\bigl[\mathbf{r}_{p_i}^B(t_m) \;-\; \mathbf{r}_{p_i}^B(t_n)\bigr]\Bigr)^{2},
\end{equation}
where $w_i$ is a weight that accounts for the local planarity quality, as discussed in Section~IV-A. The feature extraction and parameter estimation steps thus jointly solve a nonlinear least-squares problem. By iteratively refining $\mathbf{R}_L^M$ and $\mathbf{r}_L^M$, the algorithm converges to an extrinsic transformation that aligns the LiDAR points across different motor angles, as detailed in Section~IV-B.

\subsection{Calibration Primitives Extraction}
\subsubsection{Kernel Selection}

For on-site fast calibration, LiMo-Calib employs a kernel selection strategy that extracts a compact set of planar primitives (called \emph{kernels}) from the raw lidar data, in order to subsequently extract feature. To ensure that each voxel contains a sufficient number of point
points for reliable kernel extraction and subsequent characterization.
To ensure that each voxel contains a sufficient number of points for reliable kernel extraction and subsequent characterization, we set the voxel size to 1 m. The next step of the adaptive neighborhood correspondence strategy requires the support of the selected kernel data to enhance the robustness of the calibration process.

\subsubsection{Adaptive Neighborhood Correspondence}

A central design choice in LiMo-Calib is determining an appropriate neighborhood size for local plane estimation, balancing the risk of over-smoothing large neighborhoods against the increased noise sensitivity of smaller ones. We employ a \emph{dynamic} $k$-nearest-neighbor (kNN) strategy to determine the local neighborhood for plane estimation. Specifically, for each voxel kernel $\mathbf{p}_c$ in the downsampled cloud, we choose
\begin{equation}\label{eq:kneighbors}
    k \;=\; \min\Bigl(k_{\max}, \;\bigl\lfloor \gamma \cdot |\mathcal{P}|\bigr\rfloor\Bigr),
\end{equation}
where $|\mathcal{P}|$ is the total number of points, $k_{\max} = 50$ is an upper bound on neighbors, and $\gamma$ is a small fraction (e.g., $1/100$). $\gamma$ can be regarded as a neighborhood scale factor which is used to establish a reasonable ratio between the global point number and the local neighborhood size, so as to obtain a more stable planar fitting result under various density distributions. This adaptive approach ensures that denser regions do not become oversmoothed by an excessively large neighborhood, while sparser regions still gather enough points for robust plane fitting. Adaptive use of voxel's information can reduce computational resources and increase the speed of the algorithm.

\subsubsection{Plane Primitives Filter and Reweight}
After the adaptive neighborhood correspondence step, LiMo-Calib refines the candidate planar primitives by applying a two-level weighting scheme to both enhance the local plane estimation and to select the most reliable features for global calibration.

\textit {Distance Weighting for Local Plane Fitting:}
During local plane estimation, each point in the $k$-nearest-neighbor (kNN) set is assigned a weight based on its distance from the kernel center. Let $d_i$ be the squared distance from the $i$-th neighbor to the kernel center, and let $d_{\max}$ be the maximum squared distance within the kNN set. The distance weight is computed as:
\begin{equation}\label{eq:distance_weight}
    w_{\text{dist},i} \;=\; 1 \;-\; \sqrt{\frac{d_i}{d_{\max}}}.
\end{equation}
This weighting reduces the influence of points that are farther away, often more prone to noise or geometric ambiguity, thereby yielding a more robust weighted covariance matrix for subsequent singular value decomposition (SVD) and normal estimation.

\textit {Planarity Weighting for Global Calibration:}
Once the local plane normal is estimated, a planarity metric is calculated to assess the quality of the planar fit. Given the singular values $\sigma_0 \geq \sigma_1 \geq \sigma_2$ of the weighted covariance matrix, the planarity is defined as:
\begin{equation}\label{eq:planarity_metric}
    \alpha \;=\; \frac{2\bigl(\sigma_1 - \sigma_2\bigr)}{\sigma_0 + \sigma_1 + \sigma_2}.
\end{equation}
Candidate planes with $\alpha$ below a threshold (e.g., 0.5) or with a high condition number $\sigma_0/\sigma_2$ (e.g., above 100) are discarded, as they are less reliable for calibration.
The final weight assigned to each planar primitive is then set proportional to its planarity value:
\begin{equation}\label{eq:final_weight}
    w_i \;=\; \alpha_i.
\end{equation}
These weights $w_i$ are subsequently used in the global calibration cost function (Equation \eqref{eq:loss_function}) to emphasize high-confidence planar correspondences while reducing the influence of weaker ones.

In summary, by combining distance weighting to improve local plane fitting with planarity-based weighting and adaptive binning for global feature selection, LiMo-Calib effectively filters out unreliable primitives and focuses on high-quality geometric features, thereby enhancing both the robustness and efficiency of the calibration process.

\subsubsection{Plane Primitives Selection using Normal Homogenization}

\begin{figure} [ht]
    \centering
    \includegraphics[width=\linewidth]{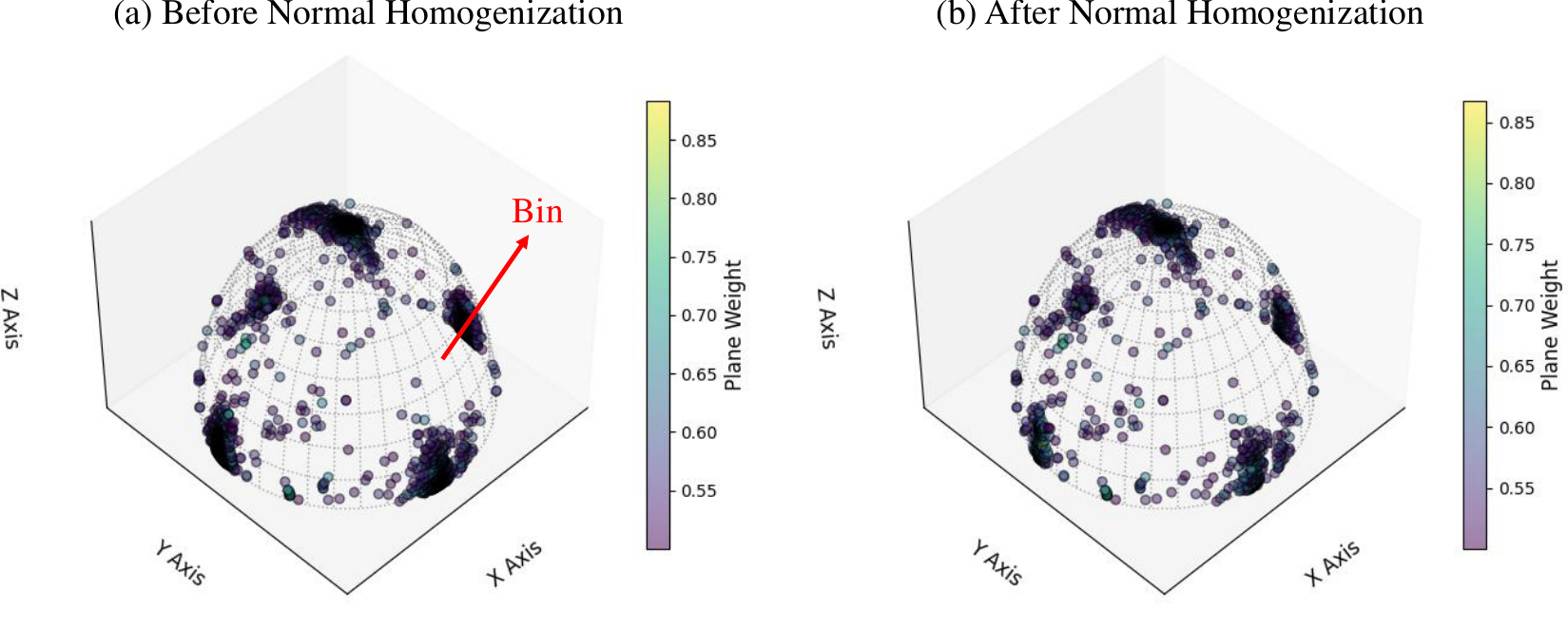}
    \caption{Plane primitives selection using normal homogenization visualized in polar coordination system. (a) A large number of raw primitives. (b) The selected primitives after normal homogenization.}
    \label{fig:bin}
\end{figure}

To ensure that the planar primitives used for calibration are uniformly distributed across different orientations, LiMo-Calib applies a normal homogenization process. In this step, each filtered candidate plane primitive has its normal vector converted into polar coordinates $(\theta, \phi)$, where
\begin{equation}
    \theta = \arccos(n_z), \quad \phi = \arctan2(n_y, n_x).
\end{equation}
The ranges of $\theta$ (from 0 to $\pi$) and $\phi$ (from $-\pi$ to $\pi$) are discretized into fixed bins (e.g., 10° intervals yielding 18 bins for $\theta$ and 36 bins for $\phi$).

Each plane primitive is assigned to a bin according to its $(\theta,\phi)$ values. To avoid over-representation of any particular orientation, if the number of primitives in a bin exceeds the average count computed over all non-empty bins, random subsampling is performed so that each bin contributes approximately the same number of primitives. The final set of primitives, along with their associated planarity weights (as defined in Eq. \eqref{eq:final_weight}), is then used in the global calibration cost function (Eq. \eqref{eq:loss_function}). An example of the selection is illustrated in Fig. \ref{fig:bin}. This normal homogenization ensures that the calibration optimization is not dominated by planes from a few over-represented orientations, thereby enhancing the efficiency and accuracy of the LiDAR-motor calibration.

\subsection{Calibration Parameter Estimation}

\subsubsection{Jacobian Estimation}

Recall the transformation (Eq. \ref{eq:point_transform}) and residual (Eq. \ref{eq:plane_factor_residual}) for the \(i\)-th planar correspondence, we have the detailed residual equation:
\begin{equation}
    \mathbf{r}_i = \mathbf{n}_i^\top \Bigl[ \mathbf{R}_M^B(t_m) \bigl(\mathbf{R}_L^M \mathbf{r}_{p_m}^L + \mathbf{r}_L^M\bigr) - \mathbf{R}_M^B(t_n) \bigl(\mathbf{R}_L^M \mathbf{r}_{p_n}^L + \mathbf{r}_L^M\bigr) \Bigr].
\end{equation} 
The deviation of $\mathbf{r}_i$ with respective to the calibration $r^M_L$ is obtained as follows:
\begin{equation}
    \partial \mathbf{r}_i /\partial \mathbf{r}^M_L = \mathbf{n}_i^\top \Bigl[ \mathbf{R}_M^B(t_m) - \mathbf{R}_M^B(t_n)\Bigr].
\end{equation} 
The deviation of $\mathbf{r}_i$ with respective to the calibration $r^M_L$ is obtained as follows:
\begin{equation}
    \partial \mathbf{r}_i /\partial \mathbf{R}^M_L = \mathbf{n}_i^\top \Bigl[ -\mathbf{R}_M^B(t_m) [\mathbf{r}_{p_m}^L]_\times + \mathbf{R}_M^B(t_n) [\mathbf{r}_{p_n}^L]_\times\Bigr],
\end{equation} 
where $[\cdot]_\times$ is the deskew matrix.

\subsubsection{Parameter Optimization}

In our optimization process, we estimate the calibration parameters $\mathbf{R}_L^M$ and $\mathbf{r}_L^M$ (along with an auxiliary time parameter) by minimizing the weighted sum of squared residuals as defined in Equation~\eqref{eq:loss_function}. Specifically, the optimization problem is formulated by constructing residual blocks for each planar correspondence using the error model in Equation~\eqref{eq:plane_factor_residual}. The parameters are divided into three parts: the rotation parameters, represented by $\mathbf{R}_L^M$, are locally parameterized using Ceres Solver to constrain updates to roll and pitch. These variables are added to the Ceres problem, and for each correspondence, a residual block is constructed with a Huber loss function applied to mitigate the effect of outliers. The Ceres solver is configured with the DENSE\_SCHUR linear solver, SUITE\_SPARSE as the sparse linear algebra library, and the Levenberg-Marquardt trust region strategy, with progress output enabled and the computation parallelized across 20 threads. After each iteration, the relative change in the final cost is computed, and the optimization terminates early if the change is below a threshold (e.g., $1 \times 10^{-6}$).

\section{Experiments}

\begin{figure} []
    \centering
    \includegraphics[width=\linewidth]{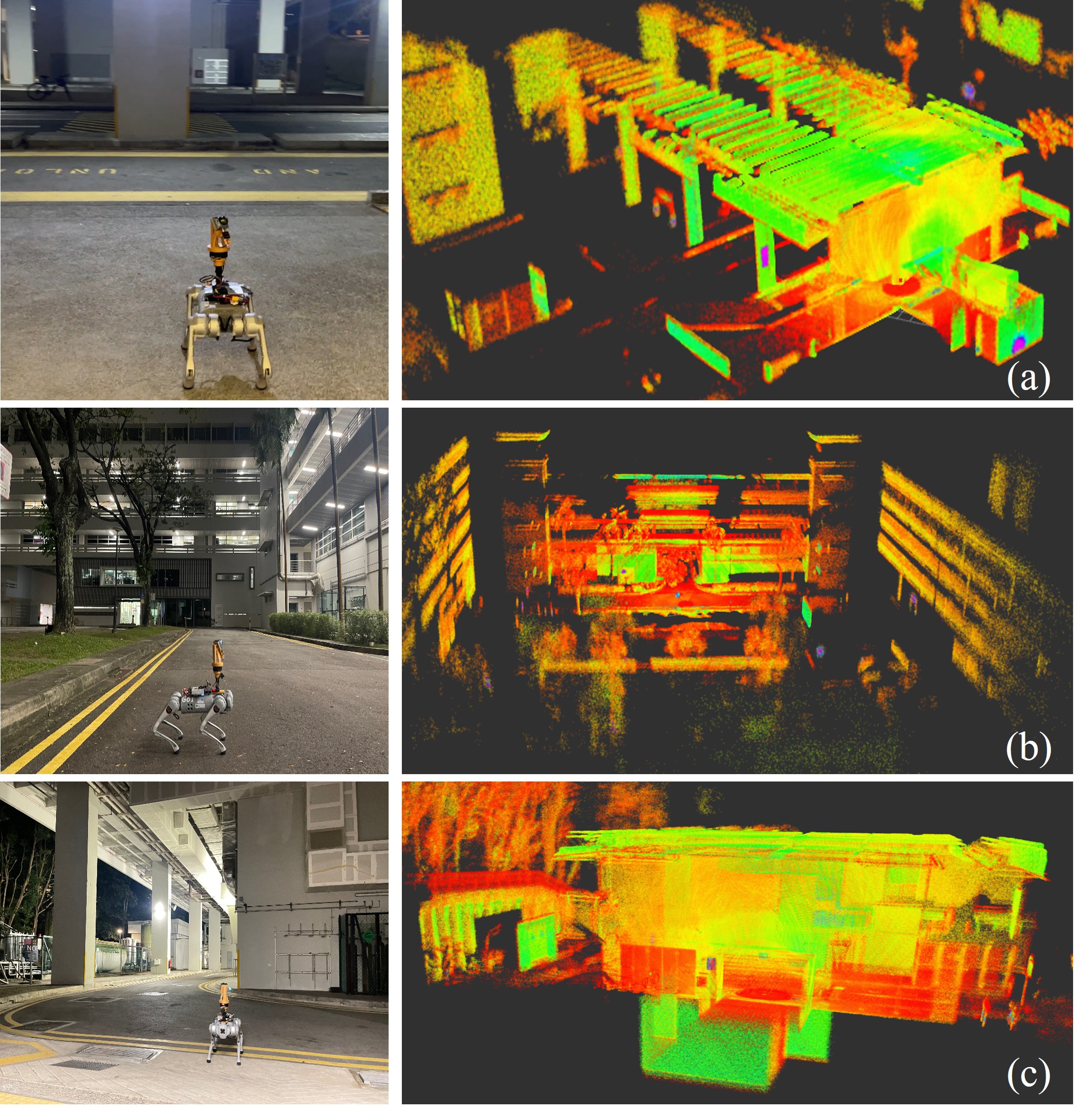}
    \caption{Overview of the three study sites (The views of the site are listed on the left side. The scanned point clouds are viewed on the right side). (a) Site a; (b) Site b; (c) Site c.}
    \label{fig:stdudy_sites}
\end{figure}

\begin{figure} []
    \centering
    \includegraphics[width=0.9\linewidth]{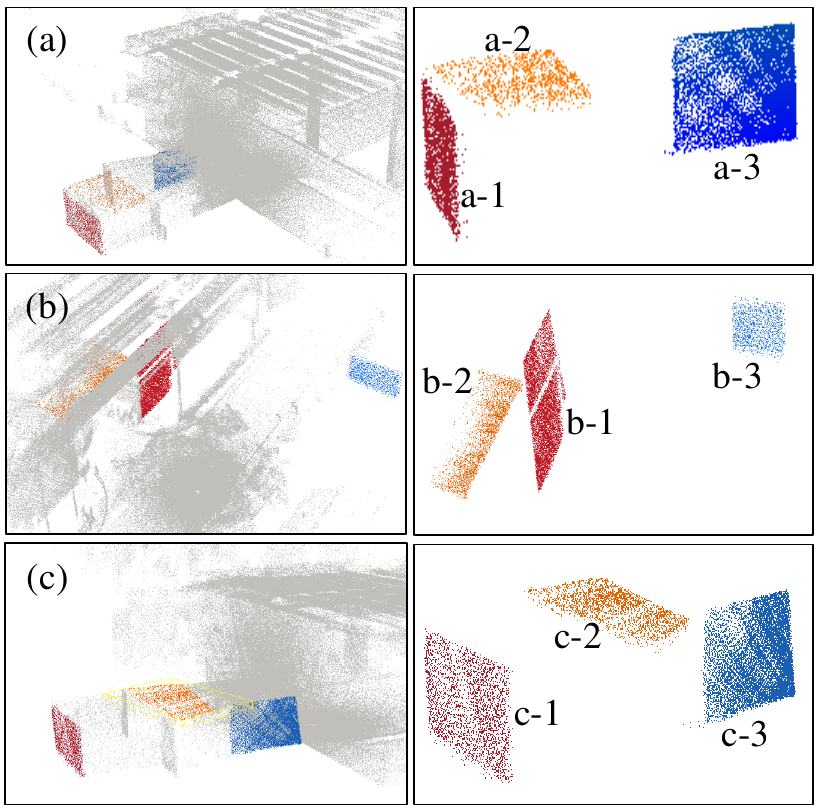}
    \caption{Calibration accuracy evaluation using planar points fitting in different sites (The selected planar points are rendered in different colors). (a) Site 1; (b) Site 2; (c) Site 3.}
    \label{fig:plane_evaluation}
\end{figure}

\subsection{Implementation Details}
LiMo-Calib is implemented in C++ and runs on Ubuntu 20.04 with ROS Noetic. The experiments were conducted on a system equipped with an Intel Core i7-10875H CPU @ 2.30 GHz.

\subsection{Evaluation Metrics}
To quantitatively assess the accuracy of the calibration, we manually select planar points from the resulting point clouds and evaluate the error of the plane alignment. The plane fitting error, is defined as the mean squared distance from the points to the plane. In our experiments, we compute the plane fitting error for the set of plane primitives extracted using both the proposed LiMo-Calib and the vanilla \cite{skaloud2006rigorous} calibration methods. The vanilla method directly optimizes the LiDAR’s extrinsic parameters by solving a nonlinear least-squares problem constructed from point cloud geometric constraints (i.e.,normal-based plane residuals). A lower error indicates a more consistent planar fit and, consequently, a higher calibration accuracy.

\subsection{Evaluation on In-house Motorized LiDAR}

\begin{figure} []
    \centering
    \includegraphics[width=0.7\linewidth]{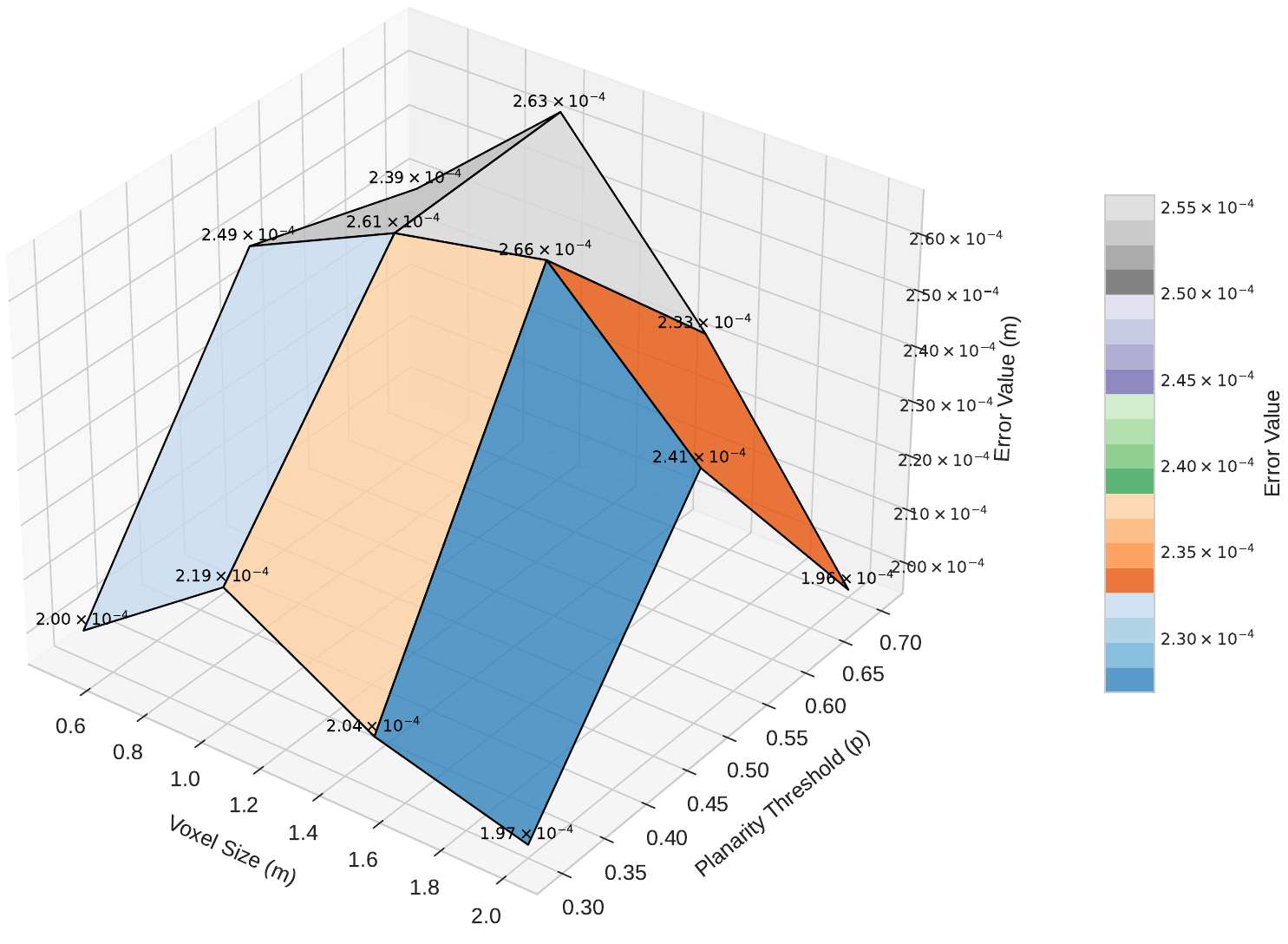}
    \caption{Error analysis using varying voxel size and planarity threshold.}
    \label{fig:pv_analysis_error}
\end{figure}

\begin{figure} []
    \centering
    \includegraphics[width=0.7\linewidth]{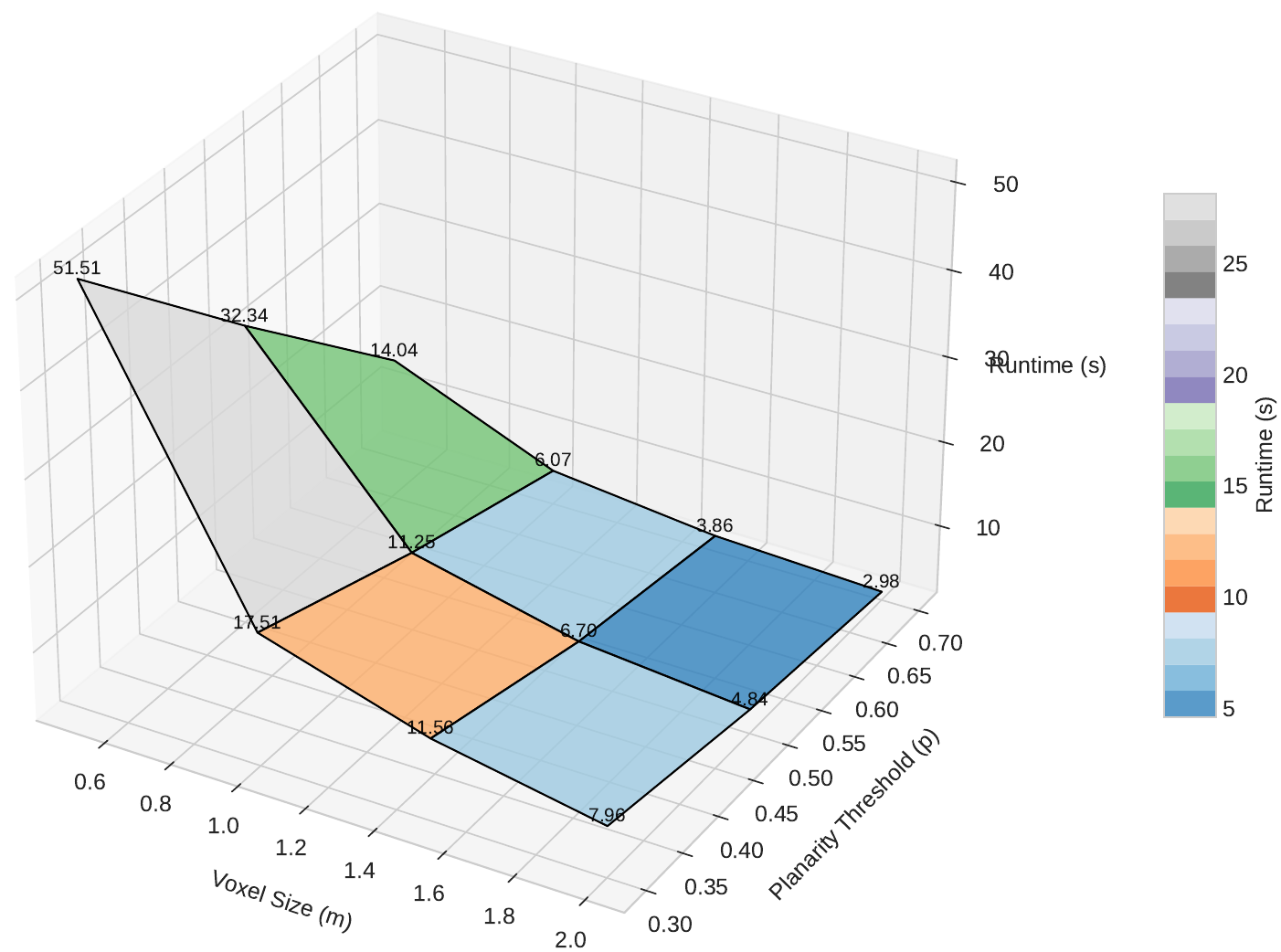}
    \caption{Runtime analysis with varying voxel Size and planarity threshold.}
    \label{fig:pv_analysis_time}
\end{figure}

\begin{table*}[]
\centering
\caption{Comparison of different methods across three sites using plane fitting errors [m].}
\label{tab:comparison}
\scriptsize
\setlength{\tabcolsep}{6pt}%
\resizebox{\linewidth}{!}{%
  \begin{tabular}{l c c c c c c c c c}
    \toprule
    & \multicolumn{3}{c}{Site 1} & \multicolumn{3}{c}{Site 2} & \multicolumn{3}{c}{Site 3} \\
    \cmidrule(lr){2-4} \cmidrule(lr){5-7} \cmidrule(lr){8-10}
    & a-1 & a-2 & a-3 & b-1 & b-2 & b-3 & c-1 & c-2 & c-3 \\
    \midrule
    Original Point Cloud & $8.61\times10^{-4}$ & $5.09\times10^{-3}$ & $1.88\times10^{-3}$ & $3.76\times10^{-3}$ & $2.03\times10^{-2}$ & $3.06\times10^{-3}$ & $5.07\times10^{-4}$ & $5.09\times10^{-3}$ & $1.78\times10^{-3}$ \\
    Vanilla(v=1.0m)              & $2.01\times10^{-4}$ & $9.02\times10^{-4}$ & $1.51\times10^{-4}$ & $1.73\times10^{-3}$ & $1.32\times10^{-3}$ & $2.33\times10^{-3}$ & $1.57\times10^{-4}$ & $1.04\times10^{-3}$ & $4.24\times10^{-4}$ \\
    LiMo-Calib(v=1.0m,p=0.5)     & $1.79\times10^{-4}$ & $4.39\times10^{-4}$ & $1.65\times10^{-4}$ & $1.66\times10^{-3}$ & $1.35\times10^{-3}$ & $1.61\times10^{-3}$ & $7.85\times10^{-5}$ & $3.51\times10^{-4}$ & $2.07\times10^{-4}$ \\
    LiMo-Calib(v=2.0m,p=0.7)     & $1.09\times10^{-4}$ & $3.61\times10^{-4}$ & $1.17\times10^{-4}$ & $1.26\times10^{-3}$ & $1.83\times10^{-3}$ & $1.49\times10^{-3}$ & $1.57\times10^{-4}$ & $3.70\times10^{-4}$ & $4.16\times10^{-4}$ \\
    \bottomrule
  \end{tabular}%
}
\end{table*}

\begin{figure} []
    \centering
    \includegraphics[width=\linewidth]{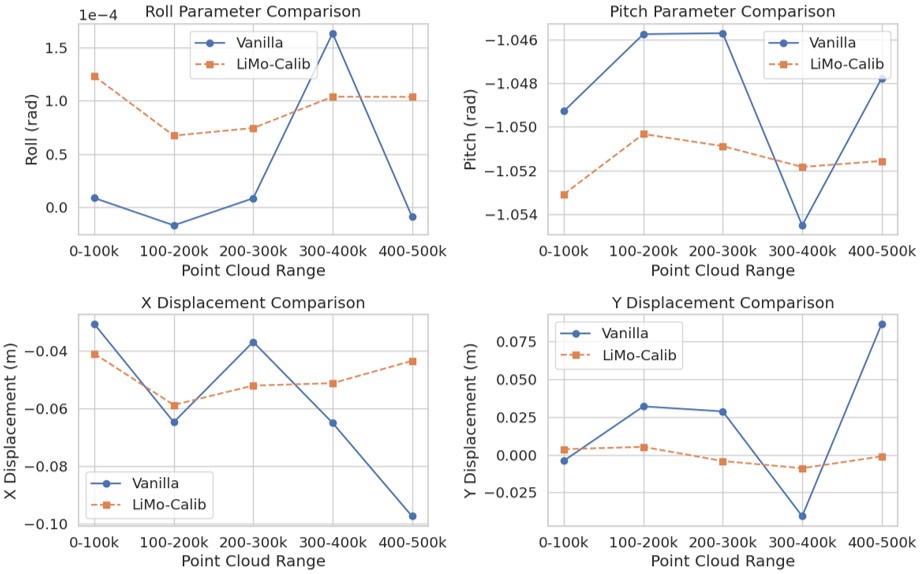}
    \caption{Stable analysis of the proposed LiMo-Calib and Vanilla method.}
    \label{fig:stable_analysis}
\end{figure}
 
To validate LiMo-Calib under real-world conditions, we conducted experiments at three representative sites in NTU campus (see Fig.~\ref{fig:stdudy_sites}). These sites differ in complexity and planar availability:

(1) \textit{Site 1} (Fig.~\ref{fig:plane_evaluation}a) contains large walls and floors, offering multiple stable planar structures.

(2) \textit{Site 2} (Fig.~\ref{fig:plane_evaluation}b) is more complex, featuring fewer discernible planes and numerous irregular objects (e.g., trees), which pose challenges for plane-based calibration.

(3) \textit{Site 3} (Fig.~\ref{fig:plane_evaluation}c) is similar to an engineering environment with pipe layouts and overhead structures.

Figure~\ref{fig:pv_analysis_error} illustrates the average plane fitting error obtained under various combinations of voxel size and planarity threshold, while Figure~\ref{fig:pv_analysis_time} presents the corresponding runtime trends for Site 1 containing 500,000 points. In this study, the voxel size parameter controls the granularity of the downsampling process. On the one hand, a larger voxel size reduces the number of points to be processed for lowering the computational burden, but may also discard fine geometric details. On the other hand, larger voxels tend to contain more points, necessitating a robust kNN strategy to extract reliable features. The planarity threshold is critical for selecting high-quality planar primitives. A stricter threshold ensures that only well-fitted planes are retained, which improves both the accuracy of the plane fitting and the overall speed by reducing the number of candidate features. The results indicate that a voxel size of 2.0\,m combined with a planarity threshold of 0.7 yields the lowest average fitting error in Site~A, achieving a favorable balance between accuracy and computational complexity. Moreover, this configuration substantially reduces runtime, thereby meeting the requirements for on-site rapid calibration.

Table~\ref{tab:comparison} compares the calibration performance across three test sites by evaluating the plane fitting errors of the uncalibrated original point cloud, the Vanilla calibration method (optimized at a voxel size of 1.0\,m), and the proposed LiMo-Calib under different parameter settings. Although the Vanilla method significantly mitigates misalignment, LiMo-Calib consistently achieves lower fitting errors, demonstrating superior accuracy and robustness. In Site~2 where the limited availability of reliable planar features due to irregular objects poses challenges. However,LiMo-Calib maintains high accuracy by effectively prioritizing high-confidence planes through its normal-based feature selection and reweighting mechanism. Similarly, in Site~3’s complex pipe layouts, LiMo-Calib adapts well to intricate geometries, resulting in tighter plane alignment. Figure~\ref{fig:stable_analysis} further presents a stability analysis comparing LiMo-Calib and the Vanilla approach over different batches of point clouds. LiMo-Calib exhibits smaller fluctuations in roll, pitch, and translation estimates, indicating a more stable calibration process.

Overall, these experimental results demonstrate that, for the present scenario, a voxel size of 2.0\,m and a planarity threshold of 0.7 not only minimize the plane fitting error but also substantially reduce runtime, thereby satisfying the demands for on-site rapid calibration. While the Vanilla method can improve calibration to some extent, its performance is consistently outperformed by LiMo-Calib in terms of both accuracy and efficiency. These findings strongly validate the efficacy of LiMo-Calib in achieving high-precision, high-efficiency calibration in complex, unstructured environments.

\subsection{Time Performance Analysis}

\begin{figure} []
    \centering
    \includegraphics[width=\linewidth]{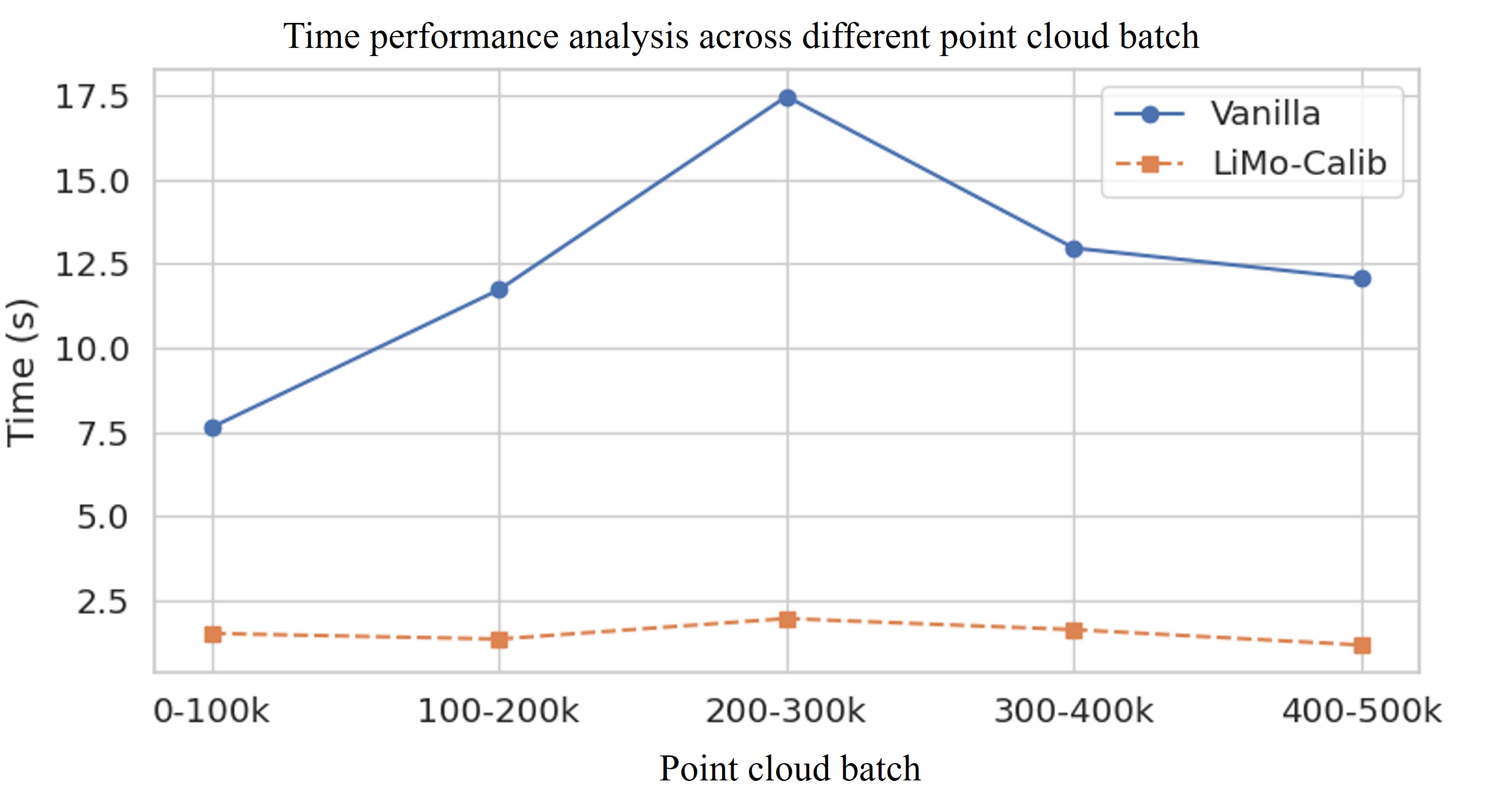}
    \caption{Time analysis of the proposed LiMo-Calib and Vanilla method.}
    \label{fig:time_analysis}
\end{figure}

\begin{figure} []
    \centering
    \includegraphics[width=\linewidth]{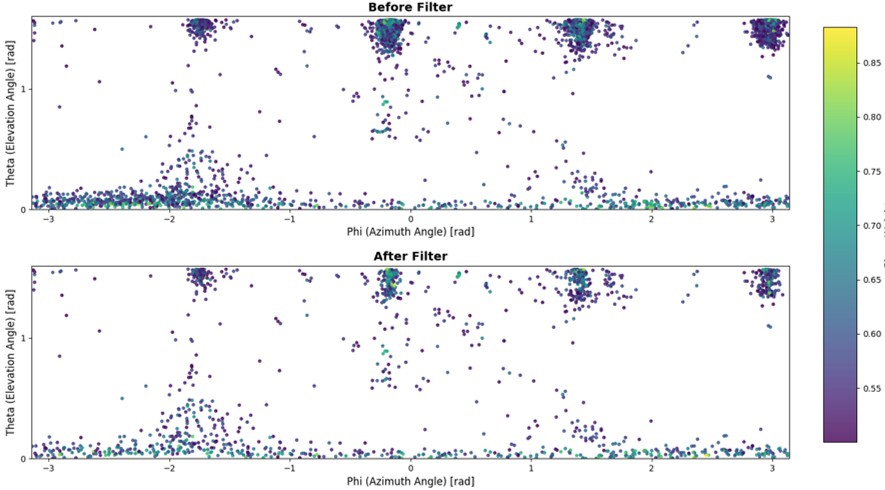}
    \caption{Normal homogenization to avoid over-represented and speed up calibration. The top shows the raw distributions of the correspondence's norm. The bottom shows the distributions of the correspondence's norm after selection by the proposed method.}
    \label{fig:normal_selection}
\end{figure}

We further evaluated time performance in the Site c containing 500,000 points, segmented into five batches of 100,000 points each. As shown in Fig.\ref{fig:time_analysis}, LiMo-Calib consistently outperforms the vanilla method in terms of convergence speed, even when processing different batches of the same scene. In addition, LiMo-Calib exhibits smaller parameter variations and lower residual errors, indicating tighter convergence and higher accuracy. A key factor behind this improvement is the normal homogenization step (Fig.\ref{fig:normal_selection}), which effectively reduces the number of planar kernels from 4964 to 1535 by eliminating over-represented correspondences with similar orientations. By focusing on an evenly distributed set of high-quality features, LiMo-Calib not only accelerates optimization but also preserves accuracy, making it particularly well-suited for on-site deployment on resource-constrained platforms.

\subsection{Evaluation of LiDAR-inertial odometry on the panoramic 3D sensing system}

\begin{figure} []
    \centering
    \includegraphics[width=\linewidth]{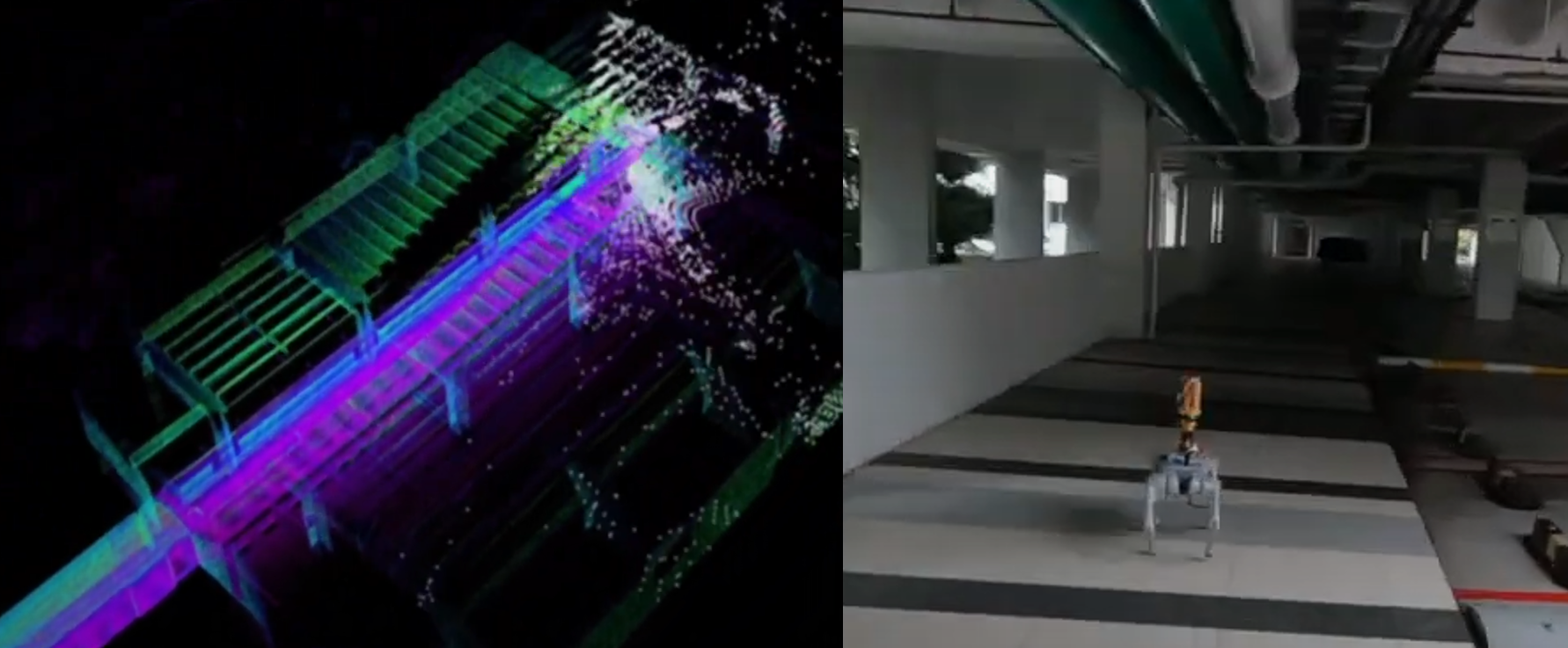}
    \caption{Mapping test using the proposed panoramic 3D sensing system.}
    \label{fig:slam}
\end{figure}

\begin{figure} []
    \centering
    \includegraphics[width=\linewidth]{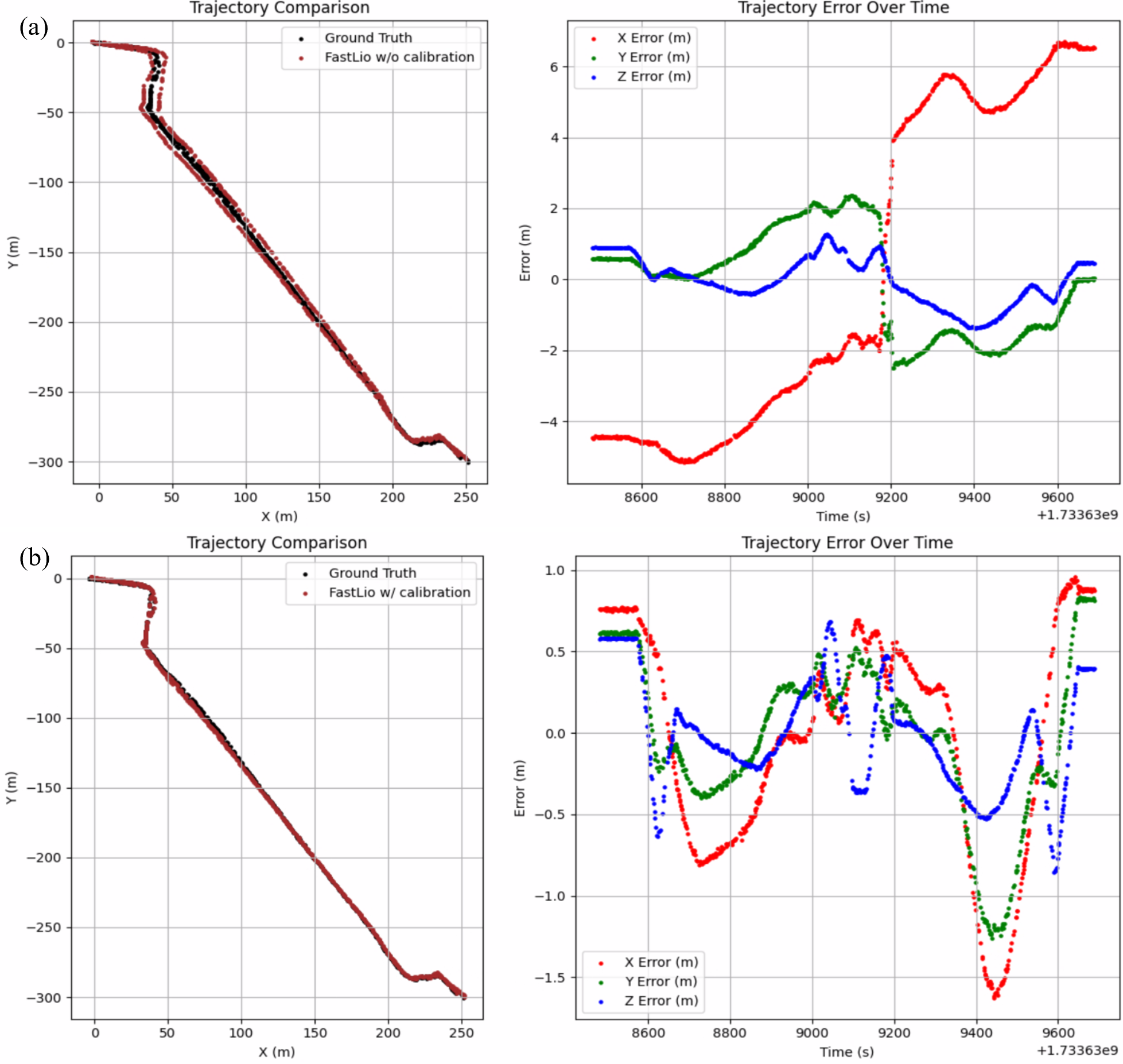}
    \caption{Absolute Position Error comparison using different calibration parameters. (a) APE without calibration. (b) APE with calibration.}
    \label{fig:ape_comparison}
\end{figure}

To further assess the performance of the proposed panoramic 3D sensing system, we conducted experiments using Fast-LIO \cite{fleishman2005robust} in a long indoor environment, as illustrated in Fig.~\ref{fig:slam}. A demonstration video is available at \url{https://www.youtube.com/watch?v=pZFQQkEqGFM}. The final Absolute Pose Error (APE) of the mapping results under different calibration settings is presented in Fig.~\ref{fig:ape_comparison}. With accurately estimated calibration parameters from the proposed LiMo-Calib, the APE is reduced to 0.09 m, whereas mapping without calibration results in an APE of 4.84 m. These results highlight the necessity of the proposed LiDAR-motor calibration method.

\section{Conclusion}
In this work, we addressed the challenge of on-site calibration for a motorized LiDAR system mounted on a quadruped robot. While motorized LiDAR significantly enhances the sensor’s field of view (FoV) and enables adaptive panoramic 3D sensing, its integration introduces calibration challenges due to high-frequency vibrations affecting the LiDAR-motor transformation. Existing methods relying on artificial targets or dense feature extraction are impractical for real-time deployment due to their computational demands. To overcome these limitations, we proposed LiMo-Calib, an efficient on-site calibration method that eliminates the need for external targets by leveraging geometric features extracted directly from raw LiDAR scans. Experimental results demonstrate that our method achieves a favorable balance between efficiency and accuracy, making it well-suited for real-world robotic applications. We will further integrate the camera and IMU into the system and investigate the related calibration tasks in the near future. 

\bibliographystyle{ieeetr}
\bibliography{ref} 
\end{document}